\documentclass[runningheads]{llncs}

\usepackage[T1]{fontenc}

\usepackage{graphicx,verbatim}
\usepackage{booktabs}
\usepackage{amsmath}
\usepackage{tabularray}
\usepackage{multicol}
\usepackage{multirow}
\usepackage[table]{xcolor}

\newcommand{\methodname}{\textit{VLEER}}
\begin{document}
\title{VLEER: Vision and Language Embeddings for Explainable Whole Slide Image Representation}
\titlerunning{Vision and Language Embeddings for Explainable WSI Representation}

\begin{comment}  %% Removed for anonymized MICCAI 2025 submission
\author{First Author\inst{1}\orcidID{0000-1111-2222-3333} \and
Second Author\inst{2,3}\orcidID{1111-2222-3333-4444} \and
Third Author\inst{3}\orcidID{2222--3333-4444-5555}}
%
\authorrunning{F. Author et al.}

% First names are abbreviated in the running head.
% If there are more than two authors, 'et al.' is used.
%
\institute{Princeton University, Princeton NJ 08544, USA \and
Springer Heidelberg, Tiergartenstr. 17, 69121 Heidelberg, Germany
\email{lncs@springer.com}\\
\url{http://www.springer.com/gp/computer-science/lncs} \and
ABC Institute, Rupert-Karls-University Heidelberg, Heidelberg, Germany\\
\email{\{abc,lncs\}@uni-heidelberg.de}}

\end{comment}

\author{Anh Tien Nguyen\inst{1} \and Keunho Byeon\inst{1} \and Kyungeun Kim\inst{2} \and  Jin Tae Kwak\inst{1}}
% index{Nguyen, TienAnh}
% index{Keunho, Byeon}
% index{Kyungeun, Kim}
% index{Kwak, JinTae}
\authorrunning{A.T. Nguyen et al.}
\institute{School of Electrical Engineering, Korea University, Seoul 02841, South Korea  \and
Seegene Medical Foundation, Seoul 04805, South Korea \\
\email{\{ngtienanh,bkh5922,jkwak\}@korea.ac.kr, kekim23@naver.com}}

\maketitle              % typeset the header of the contribution
\begin{abstract}
    Recent advances in vision-language models (VLMs) have shown remarkable potential in bridging visual and textual modalities. In computational pathology, domain-specific VLMs, which are pre-trained on extensive histopathology image-text datasets, have succeeded in various downstream tasks. However, existing research has primarily focused on the pre-training process and direct applications of VLMs on the patch level, leaving their great potential for whole slide image (WSI) applications unexplored. In this study, we hypothesize that pre-trained VLMs inherently capture informative and interpretable WSI representations through quantitative feature extraction. To validate this hypothesis, we introduce Vision and Language Embeddings for Explainable WSI Representation (\methodname), a novel method designed to leverage VLMs for WSI representation. We systematically evaluate \methodname~on three pathological WSI datasets, proving its better performance in WSI analysis compared to conventional vision features. More importantly, \methodname~offers the unique advantage of interpretability, enabling direct human-readable insights into the results by leveraging the textual modality for detailed pathology annotations, providing clear reasoning for WSI-level pathology downstream tasks.

    \keywords{Computational pathology  \and Vision-language model \and Explainability \and Whole slide image.}

\end{abstract}

\section{Introduction}
    Recently, there has been growing interest in vision-language models (VLMs), which integrate vision and language modalities by jointly learning from large-scale image-text datasets. A prominent example is CLIP \cite{clip}, which aligns visual and textual representations through contrastive learning. In computational pathology, this paradigm has been adapted to domain-specific datasets, resulting in pathology VLMs such as PLIP \cite{plip}, QUILT-Net \cite{quilt}, and CONCH \cite{conch}. These models were pre-trained on extensive histopathology datasets containing paired pathology images and descriptive textual data, effectively integrating visual and textual information in pathology. These VLMs serve as a powerful foundation for downstream tasks in computational pathology. Notably, they have achieved remarkable results in various classification tasks, such as lymph-node metastasis detection, tissue phenotyping, and Gleason grading, often without requiring further training or fine-tuning (i.e., zero-shot learning) \cite{plip,quilt,conch}.
    A close investigation of the existing research on VLMs \cite{plip,quilt,align,conch,tqx,clip} reveals that most studies have primarily focused on pre-training VLMs and their direct application to downstream tasks, overlooking two key limitations. First, most prior works mainly focus on patch-level tasks, while WSI-level applications remain largely unexplored. Second, the interpretability of textual embeddings in VLM has not been thoroughly explored, limiting their potential for providing explainable insights in computational pathology. 
            
    Herein, we hypothesize that pre-trained VLMs can inherently represent WSIs in a quantitative and interpretable manner, which can be effectively utilized for downstream tasks. To test this hypothesis, we introduce Vision and Language Embeddings for Explainable WSI Representation (\methodname), a novel approach for explainable WSI representation. We evaluate \methodname~on three WSI pathology datasets, systematically proving its effectiveness in downstream tasks. Our experimental results highlight that \methodname~not only outperforms conventional vision models in WSI analysis but also facilitates direct interpretation of results through human-readable and understandable textual representations, offering a high level of explainability in computational pathology.

\section{Methodology}
\begin{figure*}[t!]
            \centering
            \includegraphics[width=1\textwidth]{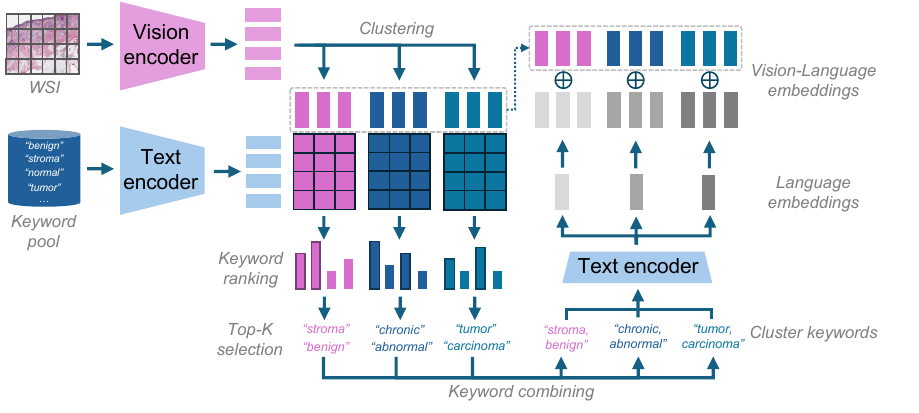}
            \caption{Vision-language embedding generation in \methodname. Tiled patches and curated keywords are embedded using a pre-trained VLM’s vision and text encoders. Clustering of vision embeddings, similarities between text and clustered vision embeddings are then measured to select the top-K keywords. These keywords are combined and used to obtain cluster-level language embeddings, which are then concatenated with the corresponding vision embeddings, forming vision-language embeddings.} \label{vleer}
        \end{figure*}

    Overall, \methodname~utilizes two components to learn explainable WSI embeddings: a task-related text pool of pathology keywords and a pre-trained pathology VLM (Fig. \ref{vleer}). 
    The text pool includes a list of pathology keywords that are relevant to downstream tasks, designed to provide explainability via human-readable and understandable pathology terms. 
    To effectively integrate visual (pathology images) and textual (pathology keywords) data, we adopt a pre-trained VLM to align these two modalities by mapping pathology patches of WSIs and their corresponding keywords, generating vision and language patch embeddings. These embeddings are then utilized for downstream analyses.
    In the following sections, we detail the construction of the text pool, the generation of vision and language embeddings, and the explainability of the proposed embeddings.
    
    \subsection{Task-related pathology text pool}       
        Inspired by~\cite{tqx}, we build a text pool of pathology-related keywords. While pathology keywords can be obtained from VLM datasets \cite{plip,conch}, these keywords are generic and not specifically tailored to downstream tasks. Accordingly, we collect task-specific keywords from relevant literature, illustrating the histology of tissues for each task. These keywords include pathological terms that are relevant to both normal and abnormal conditions. For instance, the text pool for the breast cancer subtyping task contains terms related to the histology of invasive ductal carcinoma (IDC), invasive lobular carcinoma (ILC), and normal breast tissues. All collected keywords are then reviewed and validated by a board-certified, experienced pathologist. 
    
    \subsection{Text-based WSI embedding}
    
        %\methodname~is based on the matching of pathology patches and keywords. 
        Given a WSI $\mathcal{W}$, a pre-trained VLM $\mathcal{M}$, and a task-related keyword pool $\mathcal{K}$ with $N_K$ keywords, \methodname~separately processes $\mathcal{W}$ and $\mathcal{K}$ to extract visual and textual embeddings. The two embeddings are then combined to produce a text-based WSI embedding. 
       
        \noindent\textbf{Visual embedding extraction.} $\mathcal{W}$ is tiled into a bag of patches $\mathcal{P} = \{p_i\}_{i=1}^{N_P}$ where $N_P$ is the number of extracted patches. The vision encoder of $\mathcal{M}$ transforms these patches into vision embeddings $\mathcal{V} = \{v_i\}_{i=1}^{N_P}$. 
        Typically, a multiple instance learning (MIL) aggregator is adopted to aggregate these embeddings into a single representative embedding \cite{abmil,clam,shao2021transmil,mb}. However, WSIs are inherently \textit{heterogeneous}~\cite{chan2023histopathology,levy2020spatial,wu2025leveraging}, and thus conventional approaches may fail to fully capture their complex and diverse characteristics. To address this, we cluster patches into distinct groups to improve the semantic meaning and accuracy of feature representations.
        For clustering, we employ Lloyd's $k$-Means method~\cite{kmeans} to partition $\mathcal{V}$ into $k$ clusters $\{\mathcal{V}_c\}_{c=1}^{k}$ where $\mathcal{V}_c = \{v_i\}_{i=1}^{N_P^c}$ is the $c$-th cluster, $N_P^c$ is the number of patches in $\mathcal{V}_c$, and $\sum_{c=1}^{k} N_P^c = N_P$.

        \noindent\textbf{Textual embedding extraction.} We adopt the text encoder of $\mathcal{M}$
        to embed all keywords in the pool $\mathcal{K} = \{k_j\}_{j=1}^{N_{{K}}}$ into textual embeddings $\mathcal{T} = \{t_j\}_{j=1}^{N_{{K}}}$. To enhance the robustness of these text embeddings, we employ various templates to generate diverse text prompts for each keyword, following ~\cite{lu2024avisionlanguage}. The final text embedding $t_j$ for each keyword is then obtained by averaging the embeddings derived from the various text prompts. 

        \noindent\textbf{Vision-Text alignment.} For each cluster, similarity scores are calculated between all visual and textual embeddings. Mathematically, for the $c$-th cluster, pair-wise similarity scores are obtained from embeddings in $\mathcal{V}_c$ and $\mathcal{T}$, producing $\mathcal{S}_c = \{s_{i,j} | i=1,...,N_P^c \wedge j=1,...,N_K\}$ with $s_{i,j}$ is the similarity score between visual embedding $v_i \in \mathcal{V}_c$ and textual embedding $t_j \in \mathcal{T}$. 

        \noindent\textbf{Cluster representative keywords retrieval.} Inspired by~\cite{tqx}, we rank all keywords based on their similarity scores with each patch image, aggregate these rankings across all patches within a cluster, and retrieve the most representative keywords for each cluster. 
        Formally, given a visual embedding $v_i \in \mathcal{V}_c$ and textual embeddings $\mathcal{T}$, the ranking of keywords is determined based on $\mathcal{S}_c$ as follows: $\forall (m,n)$, $r_{i,m} < r_{i,n}$ if $s_{i,m} < s_{i,n}$, with $r_{i,j} \in \{1,...,N_K\}$ is the rank of $t_j$ for $v_i$. The aggregated rank of each keyword is then calculated by its ranks across all patches in $\mathcal{V}_c$. Finally, we select the top-$M$ keywords with the highest ranks as the cluster representative keywords for the $c$-th cluster, denoted as $\mathcal{K}_c = \{k_j\}_{j=1}^{M}$.
        
        \noindent\textbf{Vision-language embedding generation.} To enhance performance and explainability, we propose integrating cluster representative keywords into the visual embeddings. This is based on the assumption that textual information not only provides semantic interpretability by a direct mapping between images and text keywords, but also enriches the extracted visual embeddings by incorporating complementary insights.
        Formally, the representative keywords $\mathcal{K}_c = \{k_j\}_{j=1}^{M}$ for the $c$-th cluster are concatenated by commas and forwarded through the text encoder of $\mathcal{M}$ to obtain the representative textual embedding $\mathcal{T}_c$. Subsequently, we concatenate $\mathcal{T}_c$ with all visual embeddings $v_i \in \mathcal{V}_c$ to produce the vision-language embeddings $\mathcal{V}^* = \{v_i^*\}_{i=1}^{N_P}$. These embeddings are then aggregated into a WSI-level embedding using a trainable MIL aggregator.

    \subsection{Explainability of vision and language embeddings}
        Utilizing both vision and language embeddings, \methodname~offers comprehensive explainability through visual and textual interpretation. Following clustering, adjacent patches within the same cluster are merged into a region of interest (RoI). Each RoI is annotated with the representative keywords. This annotation is region-specific and is generated using Vision-Language embeddings, referred to as a ReVL annotation. 
        We also generate heatmaps using the normalized attention scores from the MIL aggregator to illustrate the contribution of each RoI to the prediction. These complementary visualizations enhance model interpretability by associating key pathology patterns with their predictive significance.

\section{Experiments}
    \subsection{Datasets}
        We evaluate the effectiveness of \methodname~on cancer subtyping using three public WSI-level datasets from TCGA~\cite{tcga}.

        \sloppy\noindent\textbf{Non-small cell lung carcinoma subtyping on TCGA-NSCLC.} This dataset contains 958 WSIs, with 490 \textit{lung adenocarcinoma} (LUAD) and 468 \textit{lung squamous cell carcinoma} (LUSC) slides. The number of curated keywords for LUAD, LUSC, and normal lung histology are 23, 13, and 20, respectively.

        \noindent\textbf{Renal cell carcinoma subtyping on TCGA-RCC.} This dataset includes 922 WSIs, consisting of 519 \textit{clear cell renal cell carcinoma} (CCRCC), 294 \textit{papillary renal cell carcinoma} (PRCC), and 109 \textit{chromophobe renal cell carcinoma} (CHRCC) slides. The selected keywords for CCRCC, PRCC, CHRCC, and normal kidney histology are 18, 8, 6, and 31, respectively.
        
        \noindent\textbf{Breast invasive carcinoma subtyping on TCGA-BRCA.} This dataset has 1,033 WSIs, with 822 \textit{invasive ductal carcinoma} (IDC) and 211 \textit{invasive lobular carcinoma} (ILC) slides. The collected keywords for IDC, ILC, and normal breast histology are 27, 13, and 26, respectively

    \subsection{Implementation details}
        For quantitative evaluation, we compared vision and vision-language embeddings using four MIL aggregators: ABMIL~\cite{abmil}, CLAM-SB~\cite{clam}, CLAM-MB~\cite{clam}, and TransMIL~\cite{shao2021transmil}. Three metrics were employed, including accuracy ($Acc$), weighted F1 score ($F1$), and area under the receiver operating characteristic curve ($AUC$). Each experiment was repeated five times with different random seeds, and the average results were reported. All experiments were conducted for 20 epochs using the Adam optimizer. 
        The number of clusters and keywords was set to 5.

\section{Results and Discussion}
        \begin{table*}[t!]
            \centering
            \caption{Comparison between vision ({V}) and vision-language ({V-L}) embeddings for cancer sub-typing on three TCGA datasets with four different aggregators. \textbf{Bold} numbers indicate higher performance between two types of embeddings.}
            \resizebox{1\textwidth}{!}{
            \setlength{\tabcolsep}{0.8mm}{
            \begin{tabular} {l l ccc c ccc c ccc c ccc}
                \toprule
                \multirow{2}{*}{\textbf{Aggregator}} & \multirow{2}{*}{\textbf{Emb.}} & \multicolumn{3}{c}{\textbf{TCGA-NSCLC}} && \multicolumn{3}{c}{\textbf{TCGA-RCC}} && \multicolumn{3}{c}{\textbf{TCGA-BRCA}} && \multicolumn{3}{c}{\textbf{Average}} \\ 
                \cline{3-5}  \cline{7-9} \cline{11-13} \cline{15-17} & &
                $Acc$ & $F1$  & $AUC$ && 
                $Acc$ & $F1$  & $AUC$ &&
                $Acc$ & $F1$  & $AUC$ &&
                $Acc$ & $F1$  & $AUC$ \\            
                \toprule
                \multirow{2}{*}{\textbf{ABMIL}} & V &
                \textbf{0.9035}	& \textbf{0.9033} & \textbf{0.9739} &&
                \textbf{0.9425}	& \textbf{0.9338} & 0.9949	&&
                0.9313	& 0.9005 & 0.9707 
                && 0.9257 & 0.9126 & 0.9798 \\
                
                & V-L & 
                0.8989	&0.8988	&0.9679	&&
                0.9310	&0.9205	&\textbf{0.9961}	&&
                \textbf{0.9479}	&\textbf{0.9225}	&\textbf{0.9762} 
                && \textbf{0.9259} & \textbf{0.9139} & \textbf{0.9800}\\

                \midrule
                \multirow{2}{*}{\textbf{CLAM-SB}} & V &
                0.9012	&0.9011	&\textbf{0.9715}	&&
                \textbf{0.9402}	&\textbf{0.9282}	&0.9951	&&
                0.9271	&0.8915	&0.9649 
                && 0.9228 & 0.9069 & 0.9771 \\
                & V-L & 
                \textbf{0.9058}	&\textbf{0.9056}	&0.9696	&&
                0.9333	&0.9181	&\textbf{0.9954}	&&
                \textbf{0.9479}	&\textbf{0.9196}	&\textbf{0.9770} 
                && \textbf{0.9290} & \textbf{0.9144} & \textbf{0.9806} \\ 

                \midrule
                \multirow{2}{*}{\textbf{CLAM-MB}} & V &
                {0.8989}	&{0.8988}	&{0.9691}	&&
                \textbf{0.9333}	&\textbf{0.9191}	&0.9949	&&
                0.9292	&0.8949	&0.9650
                && 0.9205 & 0.9043 & {0.9764} \\ 
                & V-L & 
                \textbf{0.9103}	& \textbf{0.9103}	& \textbf{0.9699}	&&
                0.9287	&0.9131	&\textbf{0.9964}	&&
                \textbf{0.9479}	&\textbf{0.9196}	&\textbf{0.9780}
                && \textbf{0.9290} & \textbf{0.9143} & \textbf{0.9814}\\ 

                \midrule
                \multirow{2}{*}{\textbf{TransMIL}} & V &
                {0.8736}	&{0.8729}	&0.9603	&&
                0.9333	&\textbf{0.9247}	&0.9882	&&
                \textbf{0.9146}	&\textbf{0.8517}	&\textbf{0.9741}
                && 0.9072 & 0.8831 & 0.9759\\  
                & V-L & 
                \textbf{0.8920}	& \textbf{0.8918}	&\textbf{0.9688}	&&
                \textbf{0.9379}	&0.9236	& \textbf{0.9903}	&&
                0.9125	&0.8479	&0.9728
                && \textbf{0.9141} & \textbf{0.8878} & \textbf{0.9773}\\ 

                \bottomrule
            \end{tabular}
            }
            }
            
            \label{tab:result_classification} 
        \end{table*}

    \subsection{Quantitative evaluation of \methodname}
         Table \ref{tab:result_classification} shows the results of three classification tasks using four different MIL aggregators, comparing vision-only embedding and vision-language embedding. 
         On average, vision-language embeddings consistently achieved higher performance than vision-only embeddings across all evaluation metrics, aggregators, and datasets, with an improvement of $\sim$1\% in Acc and F1 score. However, the effect of vision-language embeddings varied among datasets and MIL aggregators. 
         For TCGA-NSCLC, vision-language embeddings showed better performance for all aggregators except ABMIL, with TransMIL showing a substantial improvement of 2.1\% in Acc and F1 score.
         Similarly, for TCGA-BRCA, vision-language embeddings outperformed vision-only embeddings, when paired with ABMIL, CLAM-SB, and CLAM-MB, with a notable gain of 2.3\% in Acc and 3.2\% in F1 score for CLAM-SB.
         Regarding TCGA-RCC, vision-language embeddings consistently obtained higher scores for AUC for all aggregators.
         These findings suggest that while vision-language embeddings enhance WSI classification performance, their effectiveness depends on the datasets and MIL aggregators.

    \subsection{Qualitative evaluation of \methodname} \label{sec:qualitative}
\begin{figure*}[t!]
            \centering
            \includegraphics[width=0.8\textwidth]{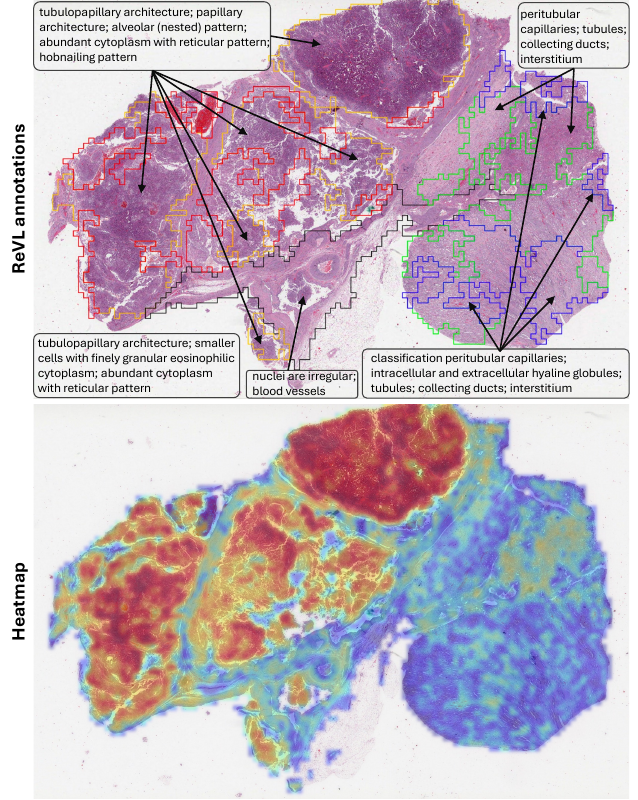}
            \caption{The ReVL annotations and attention heatmap of a papillary renal cell carcinoma in TCGA-RCC. The highly attended regions (red) in the heatmap are closely related to the patterns of papillary cancer, whereas the low attended regions (green and blue) are normal histology of renal tissues.} \label{annotation_RCC}
        \end{figure*}

\begin{figure*}[t!]
            \centering
            \includegraphics[width=0.8\textwidth]{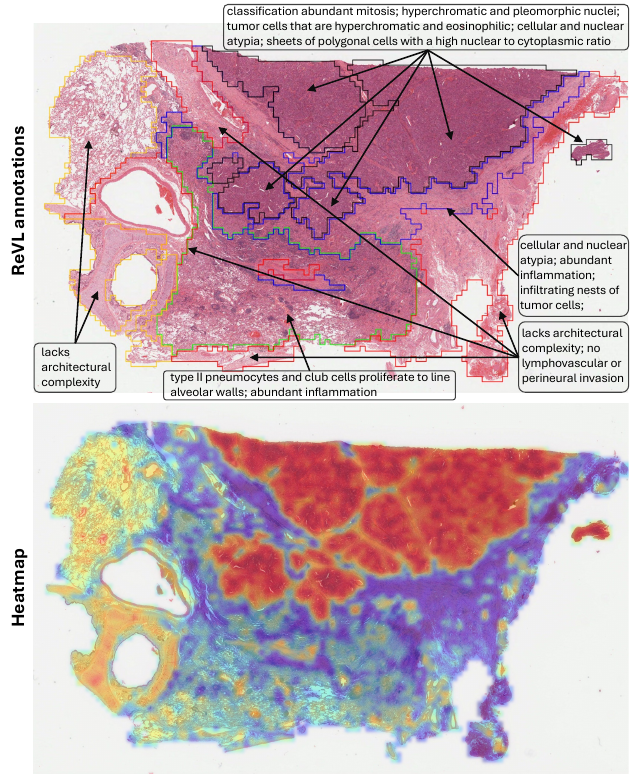}
            \caption{The ReVL annotations and attention heatmap of a lung squamous cell carcinoma in TCGA-NSCLC. The highly attended  regions (red) in the heatmap are closely related to the patterns of squamous cancer, whereas the low attended regions (green and blue) are normal histology of lung tissues.} \label{annotation_NSCLC}
        \end{figure*}

        For each WSI in the test set of three datasets (TCGA-BRCA, TCGA-NSCLC, and TCGA-RCC), we generated heatmaps and ReVL annotations for the RoIs. The ReVL annotations were reviewed and validated by a board-certified, experienced pathologist. Overall, the annotations are clinically meaningful and aligned with the histologic patterns observed in the RoIs. Fig. \ref{annotation_RCC} and Fig. \ref{annotation_NSCLC} show the heatmaps and ReVL annotations for two representative samples from TCGA-RCC and TCGA-NSCLC, respectively. As illustrated, the ReVL annotations align well with the attention heatmaps, providing a comprehensive and interpretable explanation of the model’s predictions. 
        
        Fig. \ref{annotation_RCC} presents a papillary renal cell carcinoma WSI from TCGA-RCC. The highly attended regions are annotated with \textit{abundant cytoplasm with a reticular pattern} and \textit{hobnailing pattern}, which are key pathological features of papillary carcinoma. In contrast, normal regions, annotated as \textit{peritubular capillaries} and \textit{tubules}, contribute minimally to cancer subtyping predictions. 
        Fig. \ref{annotation_NSCLC} shows a lung squamous cell carcinoma WSI from TCGA-NSCLC. The ReVL annotations capture cell-level patterns of cancerous regions, highlighting key histopathology features such as \textit{abundant mitosis} and \textit{cellular and nuclear atypia}. Non-tumor regions receive lower attention, with characteristics such as \textit{a lack of architectural complexity} and \textit{no lymphovascular or perineural invasion}. 
        In summary, the combination of ReVL annotations and heatmaps enables the precise identification of abnormal regions within a WSI along with their pathological patterns. This approach enhances explainability by linking the model’s predictions to highly attended RoIs and their corresponding pathology characteristics.

    \subsection{Explainability of \methodname}
        The level of explainability is particularly significant in computational pathology, where deep learning models often perceived as black boxes, making it difficult for clinicians and pathologists to fully trust their outputs \cite{nguyen2023gpc,tqx,song2023artificial}. Unlike traditional methods, \methodname~enhances transparency by providing text-based justifications that align with established pathology knowledge. This improved transparency and interpretability are essential for clinical adoption, as medical professionals require not only accurate predictions but also a clear rationale behind the models' decisions. By bridging the gap between machine learning and human expertise, \methodname~facilitates a more intuitive understanding of deep learning-based analyses, fostering greater confidence and acceptance in real-world diagnostic settings.

        Furthermore, this interpretability extends beyond simple classification by enabling fine-grained insights into the spatial organization of histological structures. By clustering patches and assigning pathology annotations, \methodname~preserves the contextual relationships within WSIs. This capability is particularly valuable in tasks such as tumor subtyping, cancer grading, and biomarker prediction, where localized histological variations can significantly impact clinical decisions \cite{fuchs2011computational,verghese2023computational}. Moreover, the ability to highlight RoIs alongside their corresponding pathological patterns allows pathologists to validate models' predictions.

\section{Conclusion}
    Herein, we propose \methodname~for generating vision-language embeddings in WSI representation. The method not only increases the performance on downstream tasks, but also provides explainability of the prediction. The combination of ReVL annotations and attention heatmaps forms a powerful and interpretable framework for WSI analysis. By integrating pathologically meaningful features with spatially aware visual attention, our approach enhances both model performance and transparency, making AI-driven pathology workflows more interpretable and clinically applicable by enabling clinicians to better understand models' decision-making process.
    % Future research could further refine \methodname~by incorporating more granular pathology subtypes, or leveraging hierarchical clustering for multi-scale analysis.

\bibliographystyle{splncs04}
\bibliography{references}

\end{document}